\documentclass{article}

\usepackage[nonatbib, final]{ml4ps_2023}
\usepackage[utf8]{inputenc}     
\usepackage[T1]{fontenc}        
\usepackage{microtype}          
\usepackage{booktabs}           
\usepackage{graphicx}           
\usepackage{csquotes}           
\usepackage{amsmath}            
\usepackage{amsfonts}           
\usepackage{bm}                 
\usepackage{esint}              
\usepackage{nicefrac}           
\usepackage{siunitx}            
\usepackage[dvipsnames]{xcolor} 
\usepackage[colorlinks=true, allcolors=Blue]{hyperref} 
\usepackage{bookmark}           

\usepackage[
    backend=biber,
    style=numeric-comp,
    sorting=none,
    maxcitenames=1,
    maxbibnames=2,
]{biblatex}

\DeclareFieldFormat*{title}{\enquote{#1}}
\DeclareFieldFormat*{citetitle}{\enquote{#1}}

\newcommand*{\grad}[1]{\nabla_{\!{#1}}}

\title{Score-based Data Assimilation for a\\Two-Layer Quasi-Geostrophic Model}
\author{%
    François Rozet\\
    University of Liège\\
    \texttt{francois.rozet@uliege.be}\\
    \And Gilles Louppe\\
    University of Liège\\
    \texttt{g.louppe@uliege.be}\\
}

\begin{document}

\maketitle

\begin{abstract}
    Data assimilation addresses the problem of identifying plausible state trajectories of dynamical systems given noisy or incomplete observations. In geosciences, it presents challenges due to the high-dimensionality of geophysical dynamical systems, often exceeding millions of dimensions. This work assesses the scalability of score-based data assimilation (SDA), a novel data assimilation method, in the context of such systems. We propose modifications to the score network architecture aimed at significantly reducing memory consumption and execution time. We demonstrate promising results for a two-layer quasi-geostrophic model. The code for all experiments is made available at \url{https://github.com/francois-rozet/sda}.
\end{abstract}

\section{Introduction}

$N$-layer quasi-geostrophic (NLQG) models \cite{cotter2020data} are special cases of the Navier-Stokes equations that have been extensively used to describe the dynamics of oceans and atmospheres. NLQG models consider a stratified fluid of $N$ superimposed layers of constant uniform densities; the layers being stacked according to increasing density. The state of the fluid is fully described by the potential vorticity field $q$ and the velocity fields $(u, v)$ of each layer. In practice, it is necessary to discretize temporally and spatially to solve/integrate the quasi-geostrophic equations, which leads to physical phenomena occurring at a smaller scale than the numerical resolution to be missed \cite{cotter2020data, bolton2019applications, ross2023benchmarking, mangeleer2023ocean}. In the long run, neglecting such phenomena can lead to poor simulations. It is therefore common to simulate such models at high temporal and spatial resolutions to ensure the quality of the simulations, but this comes at a significant computational cost.

Although they make (shallow water) assumptions to neglect some terms of the Navier-Stokes equations, NLGQ models effectively capture the characteristic features of turbulent ocean systems, such as jet streams, mesoscale eddies, and ocean currents. As such, they are good candidates for designing and benchmarking data assimilation (DA) \cite{lorenc1986analysis, le1986variational, tremolet2005accounting, tremolet2007model, fisher2011weak, carrassi2018data, ecmwf2020data} algorithms. Formally, DA targets the problem of inferring the posterior distribution
\begin{equation}
    p(x_{1:L} \mid y) = \frac{p(y \mid x_{1:L})}{p(y)} p(x_1) \prod_{i = 1}^{L - 1} p(x_{i+1} \mid x_i)
\end{equation}
of discrete-time state trajectories $x_{1:L} = (x_1, x_2, \dots, x_L)$ given an observation $y$ resulting from an observation process $p(y \mid x_{1:L})$. In geosciences, the physical model underlying transition dynamics $p(x_{i+1} \mid x_i)$ is typically well known and the observation process is generally formulated as $y = \mathcal{A}(x_{1:L}) + \eta$, where $\mathcal{A}$ is the measurement function and $\eta$ is a stochastic additive term that accounts for instrumental noise and systematic uncertainties.

In this setting, the simulation quality is very important as it strongly impacts the relevance of inference results. Unfortunately, high resolutions rapidly become a computational burden for classical DA algorithms, such as variational methods \cite{lorenc1986analysis, le1986variational, tremolet2005accounting, tremolet2007model, fisher2011weak} that require to repeatedly differentiate through the physical model, and Monte Carlo methods \cite{evensen1994sequential, liu1998sequential, hamill2006ensemble, doucet2011tutorial} that conduct large numbers of simulations.

Conversely, score-based data assimilation (SDA) \cite{rozet2023score}, a novel DA method drawing its roots from score-based generative modeling, only relies on the physical model to generate training data. After training, inference can be carried out without relying on the model and at lower temporal and spatial resolutions. However, applying SDA to (very) high-dimensional systems remains an engineering challenge, which we attempt to address in this work.

\section{Background}

Score-based generative modeling has shown remarkable capabilities, powering many of the latest advances in image, video and audio generation \cite{rombach2022high, ramesh2022hierarchical, saharia2022photorealistic, ho2022video, goel2022raw}. In this section, we shortly review score-based generative models and outline how they can be used for solving data assimilation problems with SDA \cite{rozet2023score}.

\paragraph{Continuous-time score-based generative models}
Adopting the notations of \textcite{rozet2023score}, samples $x$ drawn from a distribution $p(x)$ are progressively perturbed through a continuous-time stochastic diffusion process. This process determines a series of marginal distributions $p(x(t))$ from $t=0$ to $t=1$ for which $p(x(0)) \approx p(x)$ (clean data) and $p(x(1)) \approx \mathcal{N}(0, I)$ (pure noise). By design, the diffusion process can be \enquote{reversed} in order to transform noise $x(1) \sim \mathcal{N}(0, I)$ into data samples $x(0) \sim p(x(0))$. However, this reverse process requires access to the quantity $\grad{x(t)} \log p(x(t))$ known as the score of $p(x(t))$.

A score network is a neural network $s_\phi(x(t), t)$ trained -- usually by denoising score matching \cite{hyvarinen2005estimation, vincent2011connection} -- to approximate the score $\grad{x(t)} \log p(x(t))$. After training, a score network can be plugged into the reverse process as a substitute for the true score in order to generate data samples from $p(x(0))$. A handy property of score-based models is that, under some assumptions on the likelihood $p(y \mid x)$, it is possible to approximate the posterior score
\begin{equation}
    \grad{x(t)} \log p(x(t) \mid y) = \grad{x(t)} \log p(x(t)) + \grad{x(t)} \log p(y \mid x(t))
\end{equation}
without retraining the score network $s_\phi(x(t), t)$, and hence to generate data samples from $p(x(0) \mid y)$.

\paragraph{Score-based data assimilation}
In score-based data assimilation, the data we try to model are trajectories $x_{1:L}$ and we therefore approximate the trajectory score $\grad{x_{1:L}(t)} \log p(x_{1:L}(t))$. \textcite{rozet2023score} show that an element $\grad{x_{i}(t)} \log p(x_{1:L}(t))$ of the trajectory score can be approximated locally by $\grad{x_{i}(t)} \log p(x_{i-k:i+k}(t))$ for some $k \geq 1$. They therefore propose to train a local score network $s_\phi(x_{i-k:i+k}(t), t)$ to approximate the score over short segments, and compose this local score at inference to generate arbitrarily long trajectories. In addition, the authors introduce a novel approximation for the likelihood score $\grad{x_{1:L}(t)} \log p(y \mid x_{1:L}(t))$ in the assumption of a Gaussian observation process $p(y \mid x_{1:L}) = \mathcal{N}(y \mid \mathcal{A}(x_{1:L}), \Sigma_y)$, which covers many observation scenarios. Finally, to prevent approximation errors from accumulating along the simulation of the reverse process, \textcite{rozet2023score} perform a few Langevin Monte Carlo corrections between each step of the discretized reverse process.

\section{Task}

Although the Kolmogorov system presented in the experiments conducted by \textcite{rozet2023score} is high-dimensional (tens of thousands of dimensions) compared to what is approachable with classical posterior inference methods, it remains small in comparison to the millions of dimensions of some operational DA systems.

In this study, we attempt to perform data assimilation for a two-layer quasi-geostrophic (2LQG) model. We choose a torus-like \qty{1000}{\kilo\meter} periodic domain, discretized into $256 \times 256$ spatial bins. The state is described by the potential vorticity field $q^l$ and the zonal and meridional velocity fields $u^l$ and $v^l$ for each layer $l \in \{1, 2\}$. A state snapshot $x = (q^1, u^1, v^1, q^2, u^2, v^2)$ is therefore a $256 \times 256$ grid with 6 channels, which exceeds the dimensionality of \textcite{rozet2023score}'s Kolmogorov system by a factor 48.

We use the \texttt{pyqg} \cite{pyqg} Python library to solve the quasi-geostrophic equations using pseudo-spectral methods. To ensure the quality of the simulation, the domain is discretized into $512 \times 512$ spatial bins and the integration timestep is set to \qty{15}{min}. We coarsen the simulation to the target resolution ($256 \times 256$) afterwards and separate two snapshots $x_i$ and $x_{i + 1}$ by a day (\qty{24}{h}). To ensure that the system has reached statistical stationarity, that is $p(x_i) = p(x_{i+1})$, it is simulated for 5 years starting from random initial conditions before saving the first state $x_1$ \cite{bolton2019applications, ross2023benchmarking, mangeleer2023ocean}.

\section{Architecture} \label{sec:architecture}

The high-dimensionality of the task at hand (and operational DA systems) introduces many engineering challenges. Notably, naively scaling the U-Net \cite{ronneberger2015unet} inspired score network architecture proposed by \textcite{rozet2023score} to our task results in memory consumption exceeding the total memory of a A5000 GPU (\qty{24}{\giga B}) when training on segments, which gets worse at inference on full trajectories. We propose several modifications to the architecture to address these problems, avoiding low-level tricks such as custom CUDA kernels.

First, at inference, except for the extremities of the trajectory, only the $k + 1$-th element of the local score network's output is actually used. The other elements are therefore wasting both memory and compute resources. Instead, we make the score network fully-convolutional in the time axis, such that it can be applied to trajectories of any length. However, we limit its temporal receptive field such that it can still be trained on short segments. To do so, we make all layers 2-D spatial convolutions (no mixing along the time axis) and strategically add a few ($2k$) 1-D temporal convolutions. Assuming the same number of channels per state, this modification roughly reduces the amount of memory consumed and computation performed at inference by a factor $2k + 1$.

Still, training on segments of $256 \times 256$ grid states consumes a lot of memory. To solve this, we combine two ideas: gradient checkpointing and inverted bottleneck blocks. Gradient checkpointing \cite{chen2016training, gruslys2016memory} is an automatic differentiation trick that consists in only keeping in memory some intermediate values of the computation graph, instead of all. The missing values are recomputed during the backward pass, effectively trading memory consumption for execution time. Inverted bottleneck blocks \cite{sandler2018mobilenetv2, liu2022convnet} consist in limiting the number of channels of a convolutional network except within residual blocks where the number of channels is expanded to perform non-linear computations and contracted back afterwards. By checkpointing only the intermediate values with a limited number of channels, we are able to reduce memory consumption by \qty{72}{\percent}, while increasing execution time by only \qty{27}{\percent}.

A diagram of the architecture incorporating the proposed modifications is provided in Figure \ref{fig:unet}.

\section{Results}

Following \textcite{rozet2023score}, we choose a variance preserving diffusion process for our experiments and simulate the reverse process in 256 evenly spaced discretization steps and one Langevin Monte Carlo correction per step. The score networks are trained once and then evaluated under various observation scenarios.

As described in Section \ref{sec:architecture}, the local score network presents a U-Net \cite{ronneberger2015unet} inspired architecture with efficiency-oriented modifications. The temporal receptive field of the score network is such that $k = 4$ and, therefore, it can be trained on segments of length $2k + 1 = 9$ instead of entire trajectories. Using \texttt{pyqg} \cite{pyqg}, we generate 1024 trajectories of 32 state snapshots, which are split into training (\qty{80}{\percent}), validation (\qty{10}{\percent}) and evaluation (\qty{10}{\percent}) sets. For practical reasons, the vorticity and velocity fields (channels) are standardized to have unit variance. We train the score network for 1024 epochs (\qty{48}{h} on a single A5000 GPU) with the AdamW \cite{loshchilov2018decoupled} optimizer and a linearly decreasing learning rate. Further architecture and training details are provided in Appendix \ref{app:details}.

\begin{figure}[!h]
    \centering
    \includegraphics[width=\textwidth]{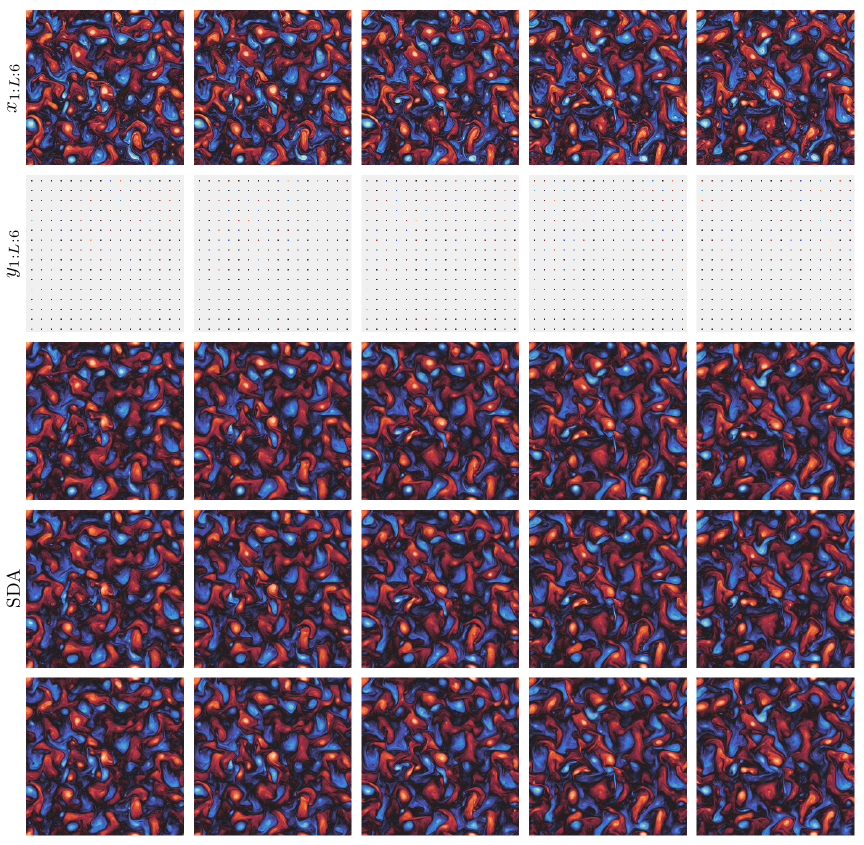}
    \caption{Example of sampled trajectories for a spatially sparse observation. States are visualized by their potential vorticity field $q$. Positive values (red) indicate clockwise rotation and negative values (blue) indicate counter-clockwise rotation. The observation $y$ corresponds to a spatial subsampling of factor $16$ of the velocity fields $(u^1, v^1, u^2, v^2)$ with small Gaussian noise ($\Sigma_y = 0.01^2 I$). SDA generates trajectories similar to the original one, despite the limited amount of information available in the observation. The three trajectories present slight physically plausible variations, as expected from sampling from a narrow posterior. We observe that the trajectories exhibit less small-scale details than the original one.}
    \label{fig:uniform}
\end{figure}

We consider a scenario where weather stations placed regularly over the domain measure the local velocities $(u^1, v^1, u^2, v^2)$ every day for a month ($L=32$). We generate an observation $y$ for a trajectory of the evaluation set. Given the observation, we sample three trajectories with SDA and find that they closely resembles the original trajectory, as illustrated in Figure \ref{fig:uniform}. A similar experiment where weather stations are placed randomly upon the domain and with more measurement noise is presented in Figure \ref{fig:random}. In both cases, however, we observe that the generated trajectories exhibit less small-scale details than a simulation, which indicates that the score network does not model the data perfectly. There are several avenues to explore in order to address this issue, such as training for longer with more data or increasing the capacity (more channels or layers) of the score network. Fourier neural operators (FNOs) could be another way to make the network more expressive and have proved to be effective for emulating partial differential equations \cite{li2020fourier, tran2023factorized, mangeleer2023ocean}.

Importantly, the three trajectories in Figure \ref{fig:uniform} were generated concurrently on a single A5000 GPU in 18 minutes and using \qty{14}{\giga B} of memory. In comparison, naively scaling the original architecture of \textcite{rozet2023score} would have required roughly 10 times more compute and 20 times more memory. With classical data assimilation algorithms, such as variational \cite{lorenc1986analysis, le1986variational, tremolet2005accounting, tremolet2007model, fisher2011weak} or Monte Carlo \cite{evensen1994sequential, liu1998sequential, hamill2006ensemble, doucet2011tutorial} methods, we estimate that inference would have required the compute equivalent of thousands of simulations, each taking roughly 3 minutes to complete using \texttt{pyqg} \cite{pyqg} on 4 CPU cores. In conclusion, we believe that SDA is ready to be tested in full operational conditions.

\begin{ack}
    François Rozet is a research fellow of the F.R.S.-FNRS (Belgium) and acknowledges its financial support.
\end{ack}


\section*{References}

\printbibliography[heading=none]

\newpage

\appendix

\section{Experiment details} \label{app:details}

\paragraph{Resources}
Experiments were conducted with the help of a cluster of GPUs. In particular, score networks were trained and evaluated concurrently, each on a single GPU with 24 GB of memory.

\begin{figure}[h]
    \centering
    \includegraphics[width=\textwidth]{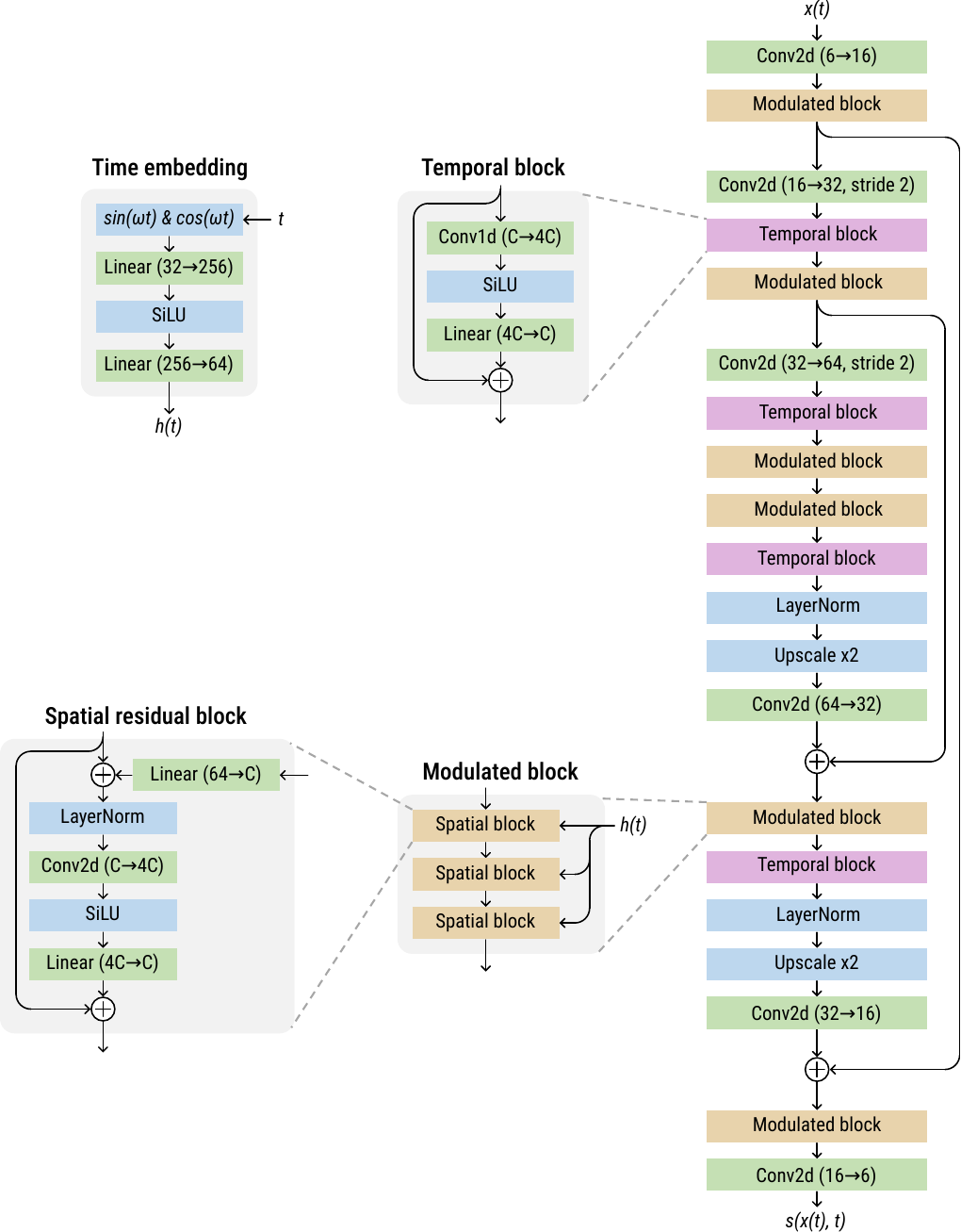}
    \caption{Schematic representation of the score network architecture. All spatial and temporal blocks are gradient checkpointed to reduce memory consumption.}
    \label{fig:unet}
\end{figure}

\paragraph{Score network}
The local score network is a U-Net \cite{ronneberger2015unet} with inverted bottleneck residual blocks \cite{he2016deep, sandler2018mobilenetv2, liu2022convnet}, SiLU \cite{elfwing2018sigmoid} activation functions and layer normalization \cite{ba2016layer}. Residual blocks either operate on the temporal axis or the spatial axes (see Section \ref{sec:architecture}). Gradient checkpointing \cite{chen2016training, gruslys2016memory} is applied to the residual blocks to reduce memory consumption. Other hyperparameters are provided in Table \ref{tab:hyperparams-2lqg} and a schematic representation of the architecture is provided in Figure \ref{fig:unet}.

\begin{table}[h]
    \centering
    \caption{Score network hyperparameters for the 2LQG experiment.}
    \label{tab:hyperparams-2lqg}
    \begin{tabular}{lcc}
        \toprule
        Spatial blocks per level & $(3, 3, 3)$ \\
        Channels per level & $(16, 32, 64)$ \\
        Inverted bottleneck factor & 4 \\
        Kernel size & 5 \\
        Padding & circular \\
        Activation & SiLU \\
        Normalization & LayerNorm \\
        \midrule
        Optimizer & AdamW \\
        Weight decay & \num{e-3} \\
        Learning rate & \num{2e-4} \\
        Scheduler & linear \\
        Epochs & 1024 \\
        Batch size & 4 \\
        \bottomrule
    \end{tabular}
\end{table}

\newpage

\newpage

\section{Assimilation examples} \label{app:example}

\begin{figure}[h]
    \centering
    \includegraphics[width=\textwidth]{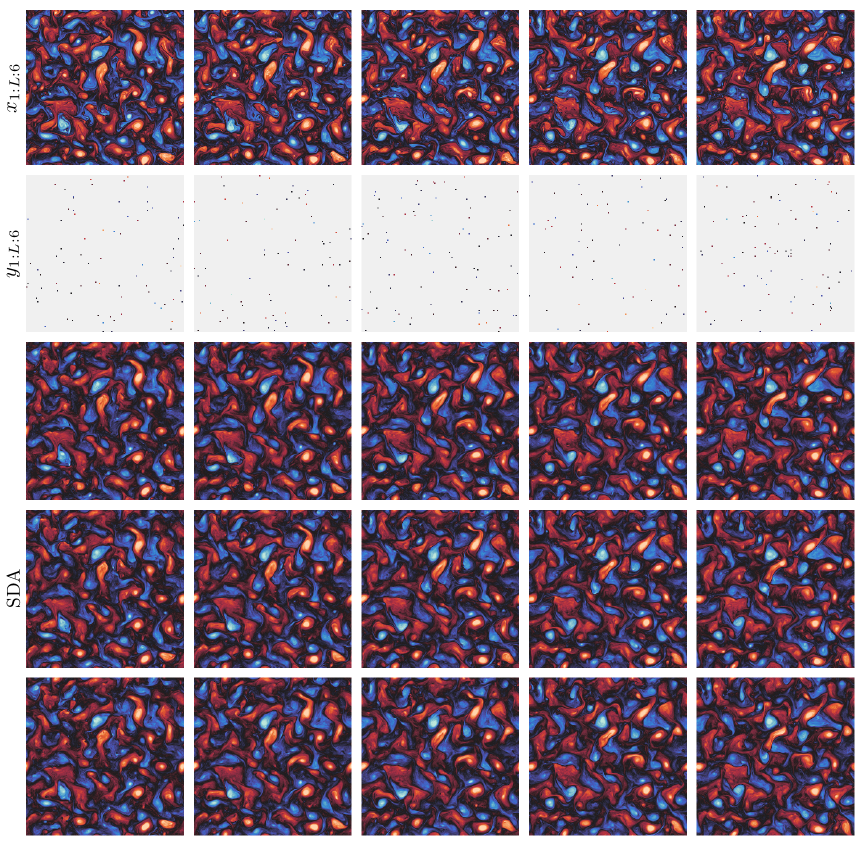}
    \caption{Example of sampled trajectories for a spatially sparse observation. The observation $y$ corresponds to a random (uniform) sampling of $\pm 80$ bins of the velocity fields $(u^1, v^1, u^2, v^2)$ with medium Gaussian noise ($\Sigma_y = 0.1^2 I$). SDA generates trajectories similar to the original one, despite the limited amount of information available in the observation. The three trajectories present slight physically plausible variations, as expected from sampling from a narrow posterior. We observe that the trajectories exhibit less small-scale details than the original one.}
    \label{fig:random}
\end{figure}

\end{document}